\documentclass[12pt]{article}
\usepackage{amssymb,graphicx}
\usepackage{amsmath}
\title{Logistic Regression as Soft Perceptron Learning}
\author{Ra\'ul Rojas}
\begin{document}

\maketitle

\begin{abstract}
We show that gradient ascent for logistic regression has a connection with the perceptron learning algorithm. Logistic learning is the ``soft'' variant of perceptron learning.
\end{abstract}

 Logistic Regression is used to build classifiers with a function which can be given a probabilistic interpretation. We are given the data set $(x_i,y_i)$, for $i=1,\ldots,N$, where the $x_i$ are $(n+1)$-dimensional {\em extended} vectors and the $y_i$ are zero or one (representing the positive or negative class, respectively).  We would like to build a function $p(x,\beta)$, which depends on a single $(n+1)$-dimensional parameter vector $\beta$ (like in linear regression) but where $p(x,\beta)$ approaches one when $x$ belongs to the positive class, and zero if not. An extended vector $x$ is one in which the collection of features has been augmented by  attaching the additional component 1 to  $n$ features, so that the scalar product of the $\beta$ and $x$ vectors can be written as
 $$
 \beta^{\rm T}x = \beta_0 + \beta_1 x _1 + \cdots + \beta_n x_n
 $$
 The extra component allows us to handle a constant term $\beta_0$ in the scalar product in an elegant way.
 
 A proposal for a function such as the one described above is
 $$
 p(x,\beta)= exp( \beta^{\rm T}x)/(1+ exp(\beta^{\rm T}x))
 $$
where $p(x,\beta)$ denotes the probability that $x$ belongs to the positive class. The function is always positive and never greater than one. It saturates asymptotically to 1 in the direction of $\beta$. Note that the probability of $x$ belonging to the negative class is given by:
$$
1- p(x,\beta)=1/(1+ exp(\beta^{\rm T}x))
$$
With this interpretation we can adjust $\beta$ so that the data has maximum likelihood. If $N_1$ is the number of data points in the positive class and $N_2$ the number o data points in the negative class, the likelihood is given by the product of all points probabilities
$$
L(\beta)= \prod^{N_1}p(x_i,\beta)\prod^{N_2}(1-p(x_i,\beta))
$$
We want to maximize the likelihood of the data, but we usually maximize the log-likelihood, since the logarithm is a monotonic function. The log-likelihood of the data is obtained taking the logarithm of $L(\beta)$. The products transform into sums of logarithms of the probabilities:
$$
\ell(\beta) = \sum^{N_1}\beta^{\rm T}x_i-\sum^{N_1}log(1+exp(\beta^{\rm T}x_i)) - \sum^{N_2}log(1+exp(\beta^{\rm T}x_i))
$$
which can be simplified to
$$
\ell(\beta) = \sum^{N}y_i\beta^{\rm T}x_i-\sum^{N}log(1+exp(\beta^{\rm T}x_i)) 
$$
where $N=N_1+N_2$ and $y_i=0$ for all points in the negative class.

If we want to find the best parameter $\beta$ we could set the gradient of $\ell(\beta)$ to zero and solve for $\beta$. However the nonlinear function makes an analytic solution very difficult. Therefore we can try to maximize $\ell(\beta)$ numerically, using gradient ascent. Therefore we compute the derivative of $\ell(\beta)$ relative to the vector $\beta$:
$$
\nabla \ell(\beta) = \sum^N (y_i x_i - (exp(\beta^{\rm T}x_i)x_i)/(1+exp(\beta^{\rm T}x_i)
$$
which reduces to
$$
\nabla \ell(\beta) = \sum^N (y_i x_i - p(x_i,\beta)x_i) = \sum^N x_i(y_i  - p(x_i,\beta))
$$
This is a very interesting expression. It essentially says that, when doing gradient ascent, the corrections to the vector $\beta$ are computed in the following way: if $x$ belongs to the positive class ($y_i=1$) but the probability $p(x_i,\beta)$ is low, we {\em add} the vector $x_i$ to $\beta$, weighted by $y_i  - p(x_i,\beta)$. And conversely: if $x$ belongs to the negative class ($y_i=0$) but the probability $p(x_i,\beta)$ is high, we {\em subtract} the vector $x_i$ from $\beta$, weighted by $|y_i  - p(x_i,\beta)|$. This is the way the perceptron learning algorithm works, without the weights. If instead of a logistic function we had a step function for assigning ``hard'' probabilities of zero and one to the data vectors, we would obtain the perceptron learning algorithm.

\end{document}